%% file: main.tex
\documentclass[conference,compsoc]{IEEEtran}

\ifCLASSOPTIONcompsoc
  \usepackage[nocompress]{cite}
\else
  \usepackage{cite}
\fi
\ifCLASSINFOpdf
\else
\fi
\usepackage{subcaption}
\usepackage{tikz}
\usetikzlibrary{shapes.geometric, arrows}
\usepackage{subcaption}

\usepackage[numbers,sort&compress]{natbib}
\usetikzlibrary{positioning}

\tikzstyle{startstop} = [rectangle, rounded corners, minimum width=2cm, minimum height=0.8cm, text centered, draw=black, fill=red!30]
\tikzstyle{process} = [rectangle, minimum width=2cm, minimum height=0.8cm, text centered, draw=black, fill=orange!30]
\tikzstyle{subprocess} = [rectangle, minimum width=2cm, minimum height=0.8cm, text centered, draw=black, fill=yellow!30]
\tikzstyle{input} = [rectangle, minimum width=2cm, minimum height=0.8cm, text centered, draw=black, fill=blue!30]
\tikzstyle{output} = [rectangle, minimum width=2cm, minimum height=0.8cm, text centered, draw=black, fill=green!30]
\tikzstyle{arrow} = [thick,->,>=stealth]

\usepackage{algorithm}
\usepackage{algpseudocode}

\begin{document}
\title{Beyond the Frame: \\Single and multiple video summarization method with user-defined length}


% \name{Vahid Ahmadi Kalkhorani$^1$,  Qingquan Zhang$^2$, and Ting Zhu$^1$}
% %The maximum number of authors in the author list is 20. If the number of contributing authors is more than this, they should be listed in a footnote or the acknowledgement section.
% \address{
%   $^{1}$Department of Computer Science and Engineering, The Ohio State University, USA \\ 
%   $^{2}$ Gies College of Business, University of Illinois at Urbana-Champaign (UIUC), Champaign, IL, USA }

% \email{\{ahmadikalkhorani.1, zhu.3445\} @osu.edu \\ qingquan@illinois.edu}

\author{
  \IEEEauthorblockN{
    Vahid Ahmadi Kalkhorani\textsuperscript{1}, Qingquan Zhang\textsuperscript{2}, Guanqun Song\textsuperscript{3}, Ting Zhu\textsuperscript{4}
  }
  \IEEEauthorblockA{
    \textsuperscript{1,3,4}Computer Science and Engineering, The Ohio State University, Columbus, OH, USA \\
    \textsuperscript{2}Gies College of Business, University of Illinois at Urbana-Champaign, Champaign, IL, USA \\
    Email: \textsuperscript{1}ahmadikalkhorani.1@osu.edu, \textsuperscript{2}qingquan@illinois.edu, \textsuperscript{3}song.2107@osu.edu, \textsuperscript{4}zhu.3445@osu.edu
  }
}


% compsoc conferences compared to traditional conferences), use this
%Telephone: (800) 555--1212, Fax: (888) 555--1212}

% \IEEEspecialpapernotice{(Invited Paper)}

% Make the title area

\maketitle

\input{0-Abstract.tex}

\input{1-Introduction.tex}

\input{2-Approach.tex} 

% \input{3-Experiments.tex}

\input{5-FutureWork}

\input{4-Conclusion.tex}

\clearpage
\bibliographystyle{IEEEtran}
\bibliography{mybib}

\end{document}

%% file: 0-Abstract.tex
\begin{abstract}

Video smmarization is a crucial method to reduce the time of videos which reduces the spent time to watch/review a long video. This apporach has became more important as the amount of publisehed video is increasing everyday. A single or multiple videos can be summarized into a relatively short video using various of techniques from multimodal audio-visual techniques, to natural language processing approaches. Audiovisual techniques may be used to recognize significant visual events and pick the most important parts, while NLP techniques can be used to evaluate the audio transcript and extract the main sentences (timestamps) and corresponding video frames from the original video.Another apporach is to use the best of both domain. Meaning that we can use audio-visual cues as well as video transcript to extract and summarize the video. In this paper, we combine a variety of NLP techniques (extractive and contect-based summarizers) with video processing techniques to convert a long video into a single relatively short video. We desing this toll in a way that user can specify the relative length of the summarized video. We have also explored ways of summarizing and concatenating multiple videos into a single short video which will help having most important concepts from the same subject in a single short video. Out approach shows that video summarizing is a difficult but significant work, with substantial potential for further research and development, and it is possible thanks to the development of NLP models.

\end{abstract}
\noindent\textbf{Index Terms}: natural language processing, summarizing, video processing

%% file: 1-Introduction.tex
\section{Introduction}

Online video production and sharing have increased dramatically in recent years, and this growth is projected to continue. Users may find it difficult to quickly and effectively locate the material they want due to the abundance of video data accessible. A useful remedy for this issue is video summary, which enables viewers to immediately scan through extensive amounts of video footage to find the most important details and relevant areas.
Video summary is the process of extracting the most important parts from a video while excluding extraneous or unimportant material. A typical video summary is made up of key images or shots that encapsulate the main points of the video. The objective is to produce a condensed video that still delivers the crucial information but is simpler and quicker to explore.
The research community has recently given video summarizing a lot of attention, and several methods have been suggested. These methods may be generally divided into two categories: \textbf{conventional} \cite{dementhon1998video, he1999auto} and \textbf{deep learning-based} \cite{yuan2019unsupervised, ji2019video, guntuboina2022video} approaches.
Traditional video summarizing techniques choose essential frames or images by manually extracting attributes like color, texture, and motion. The most crucial parts of the film are then determined by training a machine learning model using these attributes. To choose important frames or pictures, content-based, feature-based, and graph-based techniques are frequently employed. \cite{ajmal2012video}.
The visual content of the movie is examined using \textbf{content-based} approaches to pinpoint crucial shots or frames. Techniques like histogram analysis, edge detection, and motion analysis can be used for this purpose. To identify important frames or shots, feature-based algorithms combine elements including color, texture, and motion. The video is modeled using graph-based techniques, where each node corresponds to a frame or shot and the edges show the connections between them.
While traditional methods have been successful in some applications, they have limitations when dealing with complex scenes or videos with large amounts of content. Deep learning-based methods have emerged as a promising approach to video summarization, leveraging the power of neural networks to learn complex features directly from the data.
Deep learning-based video summarization methods typically use convolutional neural networks (CNNs) or recurrent neural networks (RNNs) to extract relevant features from the video frames or shots. These features are then used to select the most important frames or shots. Attention mechanisms can also be used to highlight the most salient parts of the video.
However, deep learning-based video summarization methods face several challenges. First, they require large amounts of labeled data for training, which can be difficult to obtain for some applications. Second, they struggle to capture temporal dependencies, which are crucial for understanding the context of the video. Finally, the selection of the most important frames or shots can be subjective, and the quality of the summary can vary depending on the user's preferences.

% Literature review
\subsection{Related works}
Video summarization is a process of creating a short and concise summary of a long video. It is a challenging task due to the large amount of data that needs to be processed. In recent years, there has been a lot of research in this area, and various methods have been proposed to tackle this problem. In this section, we will discuss some of the most popular video summarization methods. 

One of the most popular video summarization methods is based on deep reinforcement learning \cite{yoon2023unsupervised}. In this method, a deep neural network is trained to select the most important frames from a video. The network is trained using a reward function that encourages it to select frames that are most representative of the video. The method has been shown to be effective in summarizing videos of different genres. 

Another popular method is based on curve simplification \cite{dementhon1998video}. In this method, the video is first divided into a set of shots. Then, keyframes are extracted from each shot. The keyframes are then represented as curves, and a curve simplification algorithm is used to select the most representative curves. The method has been shown to be effective in summarizing videos of different genres. 

A low complexity video summarization method has been proposed for different types of enterprise videos such as commercial clips, talk videos, or e-meeting screen and slide share recordings \cite{dirik2014automatic}. The generated summary comprises salient shots and/or slides that are representative of the video. The method has been shown to be effective in summarizing enterprise videos. 

Another method is based on camera motion \cite{chakraborty2015adaptive}. In this method, the camera motion is analyzed to identify the most important frames. The method has been shown to be effective in summarizing videos of different genres. 

A video summarization method based on camera motion and a graph-based approach has also been proposed \cite{guironnet2007video}. In this method, the video is first divided into a set of shots, and then keyframes are extracted from each shot. The keyframes are then represented as nodes in a graph, and a graph-based algorithm is used to select the most representative nodes. The method has been shown to be effective in summarizing videos of different genres.

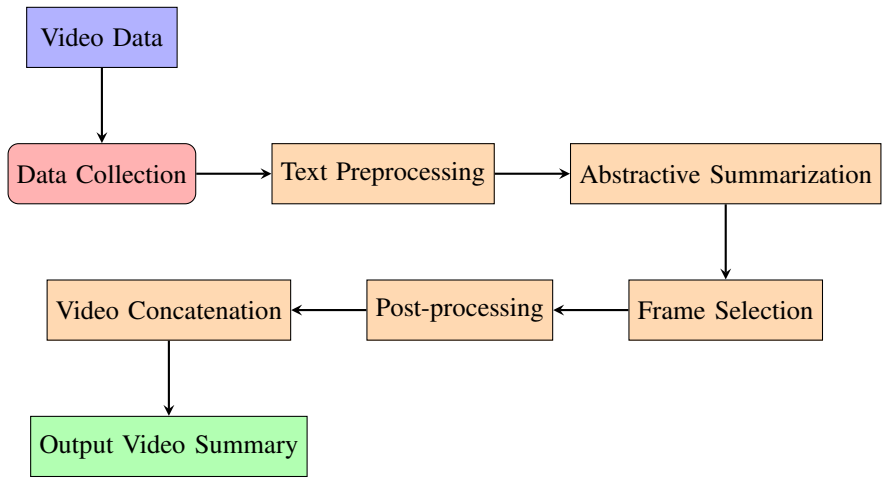
\begin{figure*}

    \centering
    \begin{subfigure}[b]{0.8\textwidth}

        \centering
        \begin{tikzpicture}[node distance=1cm and 1cm]
            \node (start) [startstop] {Data Collection};
            \node (pro1) [process, right=of start] {Text Preprocessing};
            \node (nlp) [process, right=of pro1] {Abstractive Summarization};
            \node (pro2) [process, below=of nlp] {Frame Selection};
            \node (postpro) [process, left=of pro2] {Post-processing};
            \node (concat) [process, left=of postpro] {Video Concatenation};
            \node (output) [output, below=of concat] {Output Video Summary};
            \node (input) [input, above=of start] {Video Data};
            % \node (bg1) [fill=orange!10,inner sep=0.5cm] {};
            % \node (bg2) [fill=yellow!10, inner sep=0.5cm] {};
            
            \draw [arrow] (input) -- (start);
            \draw [arrow] (start) -- (pro1);
            \draw [arrow] (pro1) -- (nlp);
            \draw [arrow] (nlp) -- (pro2);
            \draw [arrow] (pro2) -- (postpro);
            \draw [arrow] (postpro) -- (concat);
            \draw [arrow] (concat) -- (output);
        
        \end{tikzpicture}

        \label{fig:OverallMethodology}
    \end{subfigure}
    \hfill

    \caption{Summary of video summarization methodology}
\end{figure*}

%% file: 2-Approach.tex
\section{Video summarization based on text suumarizers}

\subsection{Problem Formulation}
Let $\textbf{V} = \{\textbf{v}_1, \textbf{v}_2, ..., \textbf{v}_N\}$ be a long target video clip that we want to summarize, where each frame $\textbf{v}_i$ is a RGB image matrix. The goal of video summarization is to select a subset of frames $S \subset \textbf{V}$ that best represents the video while minimizing the summary length. Formally, we can define the video summarization problem as

minimize: $\sum_{i=1}^{N} l_{v_i} I_i - P_{v_i} I_i$

subject to: $\sum_{i \in N} l_{v_i} = L $

        $l_i \in [0, L], \forall i \in \{1,2,...,N\}$
where $I_i$ is a binary variable indicating whether clip \textit{i} is selected for the summary, and $l_{v_i}$ is the length of the video clip $i$ and $P_{v_i}$ is the importance of each short clip. The objective function minimizes the weighted summary length, which is the sum of the weights of the selected frames. The constraint ensures that exactly one frame is selected for the summary, and the binary variables enforce the selection of frames.

This formulation can be extended to incorporate additional constraints or objectives, such as the maximum importance of each video clip, or the maximum length of each segment besides the total length of the video. For example, on might decide to not include video segments with importance less than a certain number which will help remove some portions of the video which are not necessary such as problem-solving, announcements by the lecturer, etc.

\subsection{Video transcript summarization}
Text summarization is a field of natural language processing that aims at generating a shorter version of a given text while retaining its most important information. As can be seen clearly, the problem definition of text summarization is very similar to video summarization and they can help each other when the other modality (text/video) is available.  In recent years, there has been a growing interest in developing text summarizer models that can automatically generate summaries for various types of texts, including news articles, scientific papers, and legal documents. In this section, we will briefly introduce different methods in this field and their potential in solving our problem i.e. video summarization.

\subsubsection{Extractive methods}
One of the earliest approaches to text summarization is the extraction-based method, which involves selecting the most important sentences or phrases from the original text and combining them to form a summary. In their survey on automatic text summarization, Hovy and Lin \cite{nazari2019survey} provide an overview of various extraction-based methods, including sentence ranking, graph-based methods, and clustering-based methods.
These methods typically involve analyzing the text to identify important sentences based on features such as sentence length, word frequency, and semantic similarity. The selected sentences are then combined to create a summary. Some popular extractive summarization methods include:
\begin{itemize}
\item Frequency-based methods, such as TF-IDF and TextRank, that rank sentences based on their frequency or importance in the text.
\item Graph-based methods, such as LexRank and SumBasic, which represent the text as a graph and use graph algorithms to identify important sentences.
\item Machine learning-based methods, such as SVM and Random Forest, which use supervised learning algorithms to classify sentences as important or unimportant.
\end{itemize}

\subsubsection{Abstractive methods}

Another popular approach to text summarization is the abstraction-based method, which involves generating a summary by paraphrasing the original text using natural language generation techniques. This method is more challenging than extraction-based methods, as it requires the model to understand the meaning of the text and generate new sentences that convey the same information. In their comprehensive study on automatic text summarization, Narayan et al. \cite{yadav2022automatic} provide an overview of various abstraction-based methods, including rule-based methods, machine learning-based methods, and deep learning-based methods.
These methods typically involve generating summaries using natural language generation techniques, such as language models and neural networks. Some popular abstractive summarization methods include:
\begin{itemize}
\item Sequence-to-sequence models, such as Encoder-Decoder and Transformer, which generate summaries using a neural network that learns to encode the input text into an embedding vector and decode the summary using this embedding vector and a decoder layer.
\item Reinforcement learning-based methods, which use a reward function to train a model to generate summaries that are both informative and readable, the reward of these models generally depends on the output text length and the importance of each sentence that appeared in the summary. This required function plays an important role in the performance and training process of these models and needs special attention and design.
\item Hybrid methods, which combine \textit{extractive} and \textit{abstractive} methods to generate summaries that are both concise and informative. These models are usually more efficient compared to abstractive methods. As they analyze a shorter text compared to the original text and yet are able to generate abstractive text which is usually more smooth compared to abstracitve results.
\end{itemize}

Although the abstractive methods generate a better representation of the long input text, the drawback for our problem (video summarization) is the inefficient process of finding corresponding timestamps of each sentence in the original text. Since the sentences in the original text do not appear directly in the summarized text, we need to compare the sentences in the output summarized text to every sentence in the original text and find the best match. On the other hand, in extractive methods, we can stop the process of finding timestamps as soon as we find a match. However, in the abstractive methods, we need to continue this process up to the end, as the best match might be at the end and next sentences. Another disadvantage of abstractive methods in our case is that these methods do not preserve the order of sentences and might change the order arbitrarily making the process of finding timestamps even harder.

\subsubsection{Hybrid methods}
\label{sec:hybrid}
Recently, there has been a growing interest in developing hybrid text summarizer models that combine both extraction-based and abstraction-based methods. These models aim to leverage the strengths of both approaches and generate more accurate and informative summaries. In their paper on text summarizer using NLP, Singh and Kaur \cite{dirik2014automatic} describe a hybrid model that uses a combination of sentence ranking and sentence compression techniques to generate summaries.

\subsection{Proposed approach}

\begin{algorithm}

\caption{Single Video Summarizer}
\label{alg:singleVideo}
\begin{algorithmic}[1]

\State Prompt the user to enter the desired summary ratio (e.g., 10\%, 20\%, etc.) and store it in a variable, \textit{ratio}
\State Use a speech-to-text model or Youtube API to convert the video's audio into text.
\State Apply NLP tool to restore any missing punctuation to the text.
\State Use an abstractive summarization model to summarize the text to the desired ratio, preserving only the most important sentences or phrases.
\ForAll{timestamp in timestamps}
\State Cut the video using the current timestamp.
\State Add the resulting video clip to a list of clips.
\EndFor

\State Concatenate the resulting video clips.
\State Output the summarized video.

\end{algorithmic}
 \end{algorithm}

\begin{figure}[h]
    \centering
    \begin{tikzpicture}[node distance=2.5cm,transform shape,scale=0.7]
        % Nodes
        \node (video_concat) [process] {Video Concatenation};
        \node (video_to_text1) [process, below of=video_concat] {Video to Text Conversion};
        \node (text_concat) [process, below of=video_to_text1] {Prepossessing};
        \node (text_summ) [process, below of=text_concat] {Text Summarization};
        \node (divide_conquer) [process, left of=text_summ, xshift=-2cm] {Divide and Conquer};
        \node (context_summ) [process, right of=text_summ, xshift=2cm] {Context-Based Summary};
        \node (video_clip_extraction) [process, below left of=text_summ, text width=3cm, xshift=-1cm] {Video Clip Extraction and Preprocessing};
        \node (video_clip_concat) [process, below of=video_clip_extraction] {Video Clip Concatenation};
        \node (cheat_sheet) [process, below right of=text_summ, xshift=1cm] {Cheat-Sheet Creation};
        % Arrows
        \draw [arrow] (video_concat) -- (video_to_text1);
        \draw [arrow] (video_to_text1) -- (text_concat);
        \draw [arrow] (text_concat) -- (text_summ);
        \draw [arrow] (text_summ) -- (divide_conquer);
        \draw [arrow] (text_summ) -- (context_summ);
        \draw [arrow] (divide_conquer) -- (text_summ);
        \draw [arrow] (context_summ) -- (text_summ);
        \draw [arrow] (text_summ) -- (video_clip_extraction);
        \draw [arrow] (text_summ) -- (cheat_sheet);
        \draw [arrow] (video_clip_extraction) -- (video_clip_concat);
        % \draw [arrow] (video_clip_concat) -- (cheat_sheet);
        % Lines
        % \draw [line] (-3,-6.5) -- (3,-6.5);
    \end{tikzpicture}
    \caption{Methodology for Summarizing and Combining Two Videos}
    \label{fig:methodology}
\end{figure}
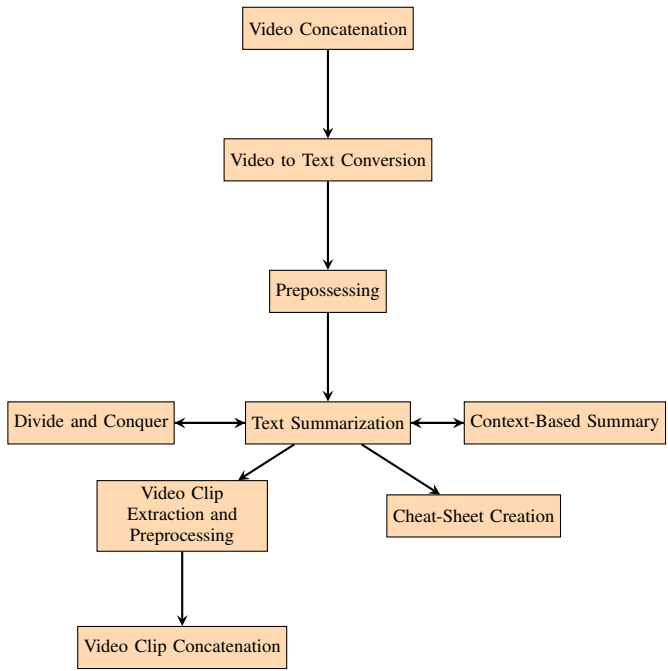

Figure \ref{fig:methodology} illustrates the overall proposed approach for video summarization. Overally, we can divide this process into 1) preprocessing, which inclues getting the data, preprocessing video and video transcript 2) processing involving summarizing video transcript, finding the corresponding timestamps in the original video, cutting the video into shorter video clips 3) postprocessing including postprocessing the summarized text for cheatsheet and abstractive summarization and concatenating the shorter video clip to obtain the final summarized video.

\subsubsection{Data collection}

In this step, we need to have access to a source of video clips (mostly lecture videos) and their transcript. Here open-source courses published on YouTube are great sources of data. More specifically, we use MIT open-course videos. YoutTube has another advantage that in the early stages of the project, we could use its API to extract the video's transcript directly without using audio-to-text models.  To this end, we send a query to youtube with the id of the video-of-interset and requests for transcripts. YouTube API returns the video transcript including the spoken sentences by the lecturer and the start and end time of each sentence in a JSON format. We also get the download link which we use to download the whole video. We also write a script that downloads the YouTube video given the download link and starting and end timestamps. This helps us reduce the time and space required to download the lecture video.

\subsubsection{Text prepossessing}
One disadvantage of YouTube transcription is that it does not contain any punctuation like ".", ",", "?", or "!", and this is of great importance as text summarizers rely heavily on text punctuation to extract sentences and their importance, this step is almost the first step of all text summarizers. To tackle this challenge, we have used a specific model trained for this purpose \cite{nagy2021automatic} namely \textit{rpunct} (\textbf{r}estore \textbf{punc}tuation). This model is a BERT model which is trained specifically to restore punctuations of a sentence. This helps us to have a clean version of the video transcript which is ready to be processed by text summarizer. The text summarizer then splits the text from this punctuation and processes each text segment. One other advantage of restoring punctuations is that we can preprocess the text and reduce the number of sentences before passing that through the summarizer. For instance, we know the questions lecturer asks from the audience has less importance compared to the answer he/she gives to the same questions. Thus, we can remove all sentences ending with a question mark.

\subsubsection{Text summarizer}
The next step in the process is to summarize the prepossessed text. To this end, we employ a pre-trained BERT model with is fine-tuned for text-summarization task \cite{zhang2019pretraining}. This model takes the prepossessed video transcript with a \textit{ratio} $\in [0,100]\%$ and ranks sentences in the text based on their importance and selects the top \textit{ratio}\% sentences. This model has the capability of being run on GPU which makes the inference time much lower. On a CPU machine (Intel Core-i7 with 8GB RAM) it takes around 1:30 to summarize a 1-hour lecture.

\subsubsection{Video Clip Extraction and pre-processing}
Given the summarized text of one video transcript, we need to find the stating and end time that, each sentence is mentioned in the video. To do so, when we are summarizing single video, we use compared every sentence in the summarized text, with sentences in the original transcript and given the fact that we have access to the starting and ending time of every single sentence in the video, we can get the timestamps of sentences in the summarized text. To make this process more efficient, we save the ending point of last found sentence and start searching from this point for the next sentence, to make sure that we are not searching the previous sentences multiple times.

In the case of multiple video summarization, since we are using an abstractive method, the exact sentence may not appear in the original text, so we need to have a comparison tool to find what portions of the video need to be cut. To solve this challenge, we compare the embedding vector of each sentence in the summarized text with the original text, and peak the highest similarity score sentence.

Then, we use \textit{ffmpeg} \cite{tomar2006converting} to extract video clips from the original video. These are usually short video clips corresponding every sentence in the summarized text. 

This process is show in Algorithm \ref{alg:singleVideo}.

\section{Diffusion video summary}

We extended the single video summarization problem to a more general and potentially more useful problem of summarizing multiple videos (lectures). Given a set of video lectures with 1-2 hours length each, we want to generate a short video clip that summarizes all these videos together into a short video clip. To final video clip will contain different segments from different lectures from the same or multiple instructors. 

\subsection{Divide and Conquer}

When summarizing multiple video we can take two approaches: 

\textbf{Approach 1}. We concatenate the transcript of all videos together and then summarize the resulting long transcript text.  In this approach, after converting audio to text, we concatenate all texts together. Then, we pass the concatenated text to the \textit{rpunc} model to restore punctuation. Then, we pass the preprocessed text to the extractive summarizer model. This model extracts the most important sentences of the joined video. In the next step we loop over all videos' texts and find the video that each sentence in the summarized text belongs to. Then, we find the corresponding starting and end timestamps and cut the video from these points and store them on disk. After cutting all clips from all videos we concatenate them together. The pseudo-code for this approach is shown in Algorithm \ref{alg:multiVideo1}. As shown in 11-20 of this pseudo-code we have nested for loop whihc is not very efficient. Moreover, lines 9 and 10 in this algorithm pass a large text to NLP models which are not computationally efficient.

\begin{algorithm}

\begin{algorithmic}[1]

\State Prompt the user to enter the desired summary ratio (e.g., 10\%, 20\%, etc.) and store it in a variable, \textit{ratio}

\ForAll{video in videos}
    \State Use a speech-to-text model or Youtube API to convert the video's audio into text and store it
\EndFor
\State Concatenate all texts 
\State Apply NLP tool to restore any missing punctuation to the text.
\State Use an extarctive summarization model to summarize the text to the desired ratio, preserving only the most important sentences or phrases, and store it in summarizedList\State Concatenate all summarized texts
\State Apply NLP tool to restore any missing punctuation to the text.
\State Use an abstractive summarization model to summarize the text with the user entered ratio

\ForAll{sentence \textbf{in} summarized text}
        
    \ForAll{video in videos}
        \ForAll{timestamp in timestamps}
        \If {sentence in video}
            \State Cut the video using the current timestamp.
            \State store video clip
        \EndIf
        \EndFor
    \EndFor
    
\EndFor

\State Concatenate the resulting video clips.
\State Output the summarized video.

\end{algorithmic}

\caption{Multi Video Summarizer - Approach 1}
\label{alg:multiVideo1} 
\end{algorithm}

\textbf{Approach 2}. We summarize each video transcript individually first. Then concatenate the summarized texts and summarized this shorter text. 

We call the second approach divide-and-conquer inspired by the famous algorithm with the same name. As can be seen, the second approach is more efficient and elegant approach compared to approach 1. First, since every video is summarized individually, in the last summarization step, we are dealing with much smaller text, which decreases computational cost. Second, in the first step, we are removing irrelevant part of each video which does not contribute to the final meaning. Finally, by applying an abstractive summarizer to the final extractive summary of the videos we can get rid of repeated concepts in the video and have a smoother video which is necessary in this case. To make the last point more clear, we need to point out that when extractive methods report important sentences in  order and sentences in the final result will always have the same order of appearance as the original text. This is not what we desire as it hurts the smoothness of the final result. Suppose the case that we have the same lecture by the same lecturer in two different semesters. If we do not use the abstractive method, in the first half of the summarized video we will see the lecture goes from start to end, and from start to end in the second part, which is obviously not desired. In this approach, we calculated the embedding similarities between each sentence in the text and picked the one with the maximum similarity score. We used a pre-trained transformer-based model to generate sentence embeddings and calculated the cosine similarity between each pair of embeddings. We then selected the sentence with the highest similarity score as the summary sentence. We repeated this process until we reached the desired summary length.

\section{Multi-modal output}

As another extension to the project we are generating text-based summaries beside the video summary which is the primary focus of this project. To do so, we are employing a hybrid approach as discussed in section \ref{sec:hybrid}. 

\subsection{Context-based summary and cheat-sheet}

To generate the text summary of the given video we employ an abstractive BERT model where we convert the whole video transcript into a single paragraph. To make this process more efficient we run the abstractive summarizer on the extractive summary of the video(s) that we generated for the video summarization (hybrid approach). This summarized paragraph can be used for searching videos among thousands of videos without the need to provide a title or keywords for each video by the publisher of the video. One example of the process is shown in Fig. \ref{fig:paragraph}.

\begin{figure}[h]
\centering
\caption{Text summary of a lecture on convolutional neural network published in MIT opencourse channel \\ \textcolor{blue}{https://www.youtube.com/watch?v=uapdILWYTzE} }
\label{fig:paragraph}
\begin{tikzpicture}
    \node[draw, rectangle, inner sep=10pt] (box) {
        \begin{minipage}{0.85\linewidth}
        The video discusses the use of neural networks to extract meaningful features from data, such as human faces, by learning a hierarchy of features that can be used to detect the presence of a face in a new image. Different filters or convolutional features are used to detect different types of features in the image. These features are then passed through fully connected layers to perform classification tasks, such as image classification. The power of this architecture extends to various tasks, such as in medicine where deep learning models can analyze medical image scans. The behavior of the neural network can be visualized using semantic segmentation maps to obtain a more detailed understanding of the image classification process.
                
        \end{minipage}
    };
\end{tikzpicture}
\end{figure}

% \subsection{Cheat-Sheet Creation}
We have also used an abstractive method to extract key-points mentioned in the video in a few bullet-points in form of a cheatsheet. One example of the process is shown in Fig. \ref{fig:cheatsheet}.

\begin{figure}[h]
\centering
\caption{Cheatsheet of a lecture on convolutional neural network published in MIT opencourse channel \\ \textcolor{blue}{https://www.youtube.com/watch?v=uapdILWYTzE} }
\label{fig:cheatsheet}
\begin{tikzpicture}
    \node[draw, rectangle, inner sep=10pt] (box) {
        \begin{minipage}{0.85\linewidth}

        \begin{itemize}
      
            \item A neural network-based approach is used to extract meaningful features from data of human faces.
            \item Filters are designed to pick up different types of features in the image.
            \item The computation of class scores is performed using convolution, non-linearity, and pooling.
            \item The architecture is powerful and extends to many tasks beyond image classification.
            \item Deep learning models can have a significant impact in medicine and healthcare.
            \item Neural networks can be visualized through semantic segmentation maps.

        \end{itemize}
                
        \end{minipage}
    };
\end{tikzpicture}
\end{figure}

\begin{algorithm}
\label{alg:multiVideo}
\caption{Multi Video Summarizer  - Approach 2}
\begin{algorithmic}[1]

\State Prompt the user to enter the desired summary ratio (e.g., 10\%, 20\%, etc.) and store it in a variable, \textit{ratio}

\ForAll{video in videos}
    \State Use a speech-to-text model or Youtube API to convert the video's audio into text.
    \State Apply NLP tool to restore any missing punctuation to the text.
    \State Use an extarctive summarization model to summarize the text to the desired ratio, preserving only the most important sentences or phrases, and store it in summarizedList
\EndFor
\State Concatenate all summarized texts
\State Apply NLP tool to restore any missing punctuation to the text.
\State Use an abstractive summarization model to summarize the text with $1/N$ ratio, where $N$ is the number of videos
\State Find starting and end time of every sentence by comparing its embedding vector with the summarized text in previous step
\ForAll{video in videos}
\ForAll{timestamp in timestamps}
\State Cut the video using the current timestamp.
\State Add the resulting video clip to a list of clips.
\EndFor
\EndFor

\State Concatenate the resulting video clips.
\State Output the summarized video.

\end{algorithmic}
 \end{algorithm}

 \section{Evaluation and Challenges}

 The evaluation of the summarized video was one of the most challenging parts. To tackle this challenge we relied on the performance of the NLP models and more specifically text summarizers. As our proposed approach highly depends on the performance of the extractive and abstractive text summarization models, the quality of output video has a one-to-one relation to the performance of these models. Thus, the quality of the output video in terms of smoothness and natural language is the same as these NLP models. In other words, our proposed procedure has the chance to be improved with the advancements in NLP models. 
 Another challenge, we faced during the project was the computational and space complexity of NLP models, especially in multi-video summarization tasks. Since our goal was to summarize tens of videos into a short video, we had to summarize tens of hours of videos (lectures) which is of course very time and computation-consuming. To tackle this problem, we proposed the divide-and-conquer approach where we summarize each video separately and then summarize these summarise again. This has multiple advantages: First, we can save the summary of the video at the upload time and save them for future usage. So, there is no need to process a video multiple times, for multiple use cases. Second, we can summarize the video in parallel in the first step instead of processing the while text sequentially.

%% file: 5-FutureWork.tex
% Future Work and Extended Considerations
\section{Future Work and Extended Considerations}

The field of video summarization presents several exciting directions for future research. One such direction involves the integration of video summarization techniques with emerging technologies in IoT and heterogeneous computing systems \cite{5,6,7,8,9,10,11,12,13,14,15,16,17}. For instance, the development of algorithms that can efficiently process and summarize video data from a multitude of IoT devices, such as security cameras and environmental sensors, is a promising area of research. This approach could leverage concurrent communication methods and heterogeneous data sources to create more comprehensive and context-aware video summaries.

The application of machine learning, particularly deep learning models, in video summarization has shown promising results \cite{34,35,36,37,38,39,40}. Future work could explore the development of models that are optimized for low-power devices \cite{22,23,24}, making video summarization more accessible in mobile and embedded systems. This could include edge devices that perform on-the-fly video summarization, aiding in real-time decision-making in various scenarios like traffic management or emergency response.

Furthermore, the integration of video summarization methods with security and privacy-preserving techniques \cite{65,66,67,68,69,70} presents a novel research area. As video data often contains sensitive information, developing summarization methods that respect user privacy and data security is critical. Techniques such as physical-layer message authentication \cite{44} and obfuscation-robust character extraction \cite{360} could be adapted to ensure the integrity and confidentiality of summarized video data.

Another intriguing area of research involves the use of video summarization in smart health and wellness applications \cite{520,530,540,550,560,570,580}. For instance, summarizing video data from fitness sessions or medical procedures can provide quick and insightful overviews for both patients and healthcare professionals. The challenge here is to develop summarization algorithms that are sensitive to the nuances and critical moments in health-related videos.

Lastly, the exploration of video summarization in sustainable energy systems and smart cities \cite{590,600,610,620} could lead to innovative applications. Summarized video data could be utilized to monitor and manage energy consumption, traffic flows, and public safety in a more efficient manner.

In conclusion, the future of video summarization is not confined to improving the algorithms' accuracy and efficiency but also extends to its application in diverse and interdisciplinary domains. The integration of video summarization with IoT, machine learning, security, and smart systems opens new pathways for research and practical applications.

%% file: 4-Conclusion.tex
\section{Conclusion}

In conclusion, in this paper, we presented a robust of effective method to summarize single and multiple videos. Our methodology provides a comprehensive framework that integrates several different techniques and processes in NLP and video processing domain, including different abstractive and context-based summaries and text preprocessors.  Our proposed method can solve two distinct problems of summarizing a single video and multiple videos at the same time. To select the most efficient way, we have designed two different processes for each problem. In the single video summarization method, we used an extractive method, while we are using a hybrid method (both extractive and abstractive) in the multi-video summarization task.  We have also generated context-based summaries which can be used for cheat-sheet generation for offline review of videos and effectively search in the context of videos without the need of adding information about the content of each video manually by the user. The presented approach has a lot of potential and usage in different use cases, such as lecture video summarization, news broadcasts, and social-media applications.